\documentclass{llncs}
\usepackage[utf8]{inputenc}
\usepackage{lmodern}
\usepackage{alltt} 
\usepackage{booktabs}
\usepackage{url}
\usepackage{tikz}
\usetikzlibrary{shapes,arrows}
\tikzstyle{line}=[draw]
\tikzstyle{rect} = [draw, rectangle,fill=white, node distance=3cm,
    minimum height=2em]
\usetikzlibrary{trees,positioning,fit}
\usepackage{pgfplots}
\usepackage{float}
\usepackage{subfig}
\usepackage{amssymb}
\usepackage{xspace}
\usepackage{amsmath}

\def\holfour{\textsf{HOL4}\xspace}
\def\isabellehol{\textsf{Isabelle/HOL}\xspace}
\def\hollight{\textsf{HOL Light}\xspace}

\def\matita{\textsf{Matita}\xspace}
\def\coq{\textsf{Coq}\xspace}

\def\eprover{\textsf{E-prover}\xspace}

\def\holyhammer{\textsf{HOL(y)Hammer}\xspace}
\def\sledgehammer{\textsf{Sledgehammer}\xspace}

\begin{document}

\title{\holfour-\hollight: Sharing proof advice through \holyhammer}
\title{Sharing HOL4 and HOL Light proof advice}
\title{Learning from the combined library of HOL4 and HOL Light}
\title{Learning and advice in combined HOL libraries}
\title{Sharing HOL4 and HOL Light proof knowledge}

\author{Thibault Gauthier \and Cezary Kaliszyk}
\institute{University of Innsbruck, Austria\\
\{thibault.gauthier,cezary.kaliszyk\}@uibk.ac.at}
\maketitle

\begin{abstract}
  New proof assistant developments often involve concepts similar to already formalized
  ones. When proving their properties, a human can often take inspiration from the
  existing formalized proofs available in other provers or libraries.  In this paper we
  propose and evaluate a number of methods, which strengthen proof automation by
  learning from proof libraries of different provers.  Certain conjectures can be
  proved directly from the dependencies induced by similar proofs in the other
  library. Even if exact correspondences are not found, learning-reasoning systems can
  make use of the association between proved theorems and their characteristics to
  predict the relevant premises. Such external help can be further combined with
  internal advice.
  We evaluate the proposed knowledge-sharing methods by reproving the HOL Light and HOL4
  standard libraries. The learning-reasoning system HOL(y)Hammer, whose single best strategy could automatically
  find proofs for 30\% of the HOL Light problems, can prove 40\% with the knowledge from HOL4.


\end{abstract}

\section{Introduction}

As \emph{Interactive Theorem Prover} (ITP) libraries were developed for decades,
today their size can often be measured in tens of thousands of facts~\cite{miningafp,mml-homepage}.
The theorem provers typically differ in their logical foundations,
interfaces, functionality, and the available formalized knowledge.
Even if the logic and the interface of the chosen prover are convenient
for a user's purpose, its library often lacks some formalizations already
present in other provers' libraries.
Her only option is then to manually repeat the proofs inside her prover. 
She will then take ideas from the previous proofs and adapt them to the specifics of her prover.
This means that in order to formalize the desired theory, the user needs to
combine the knowledge already present in the library
of her prover, with the knowledge present in the other formalization.

We propose an approach to automate this time-consuming process: 
It consists of overlaying the two libraries using concept matching 
and using learning-assisted automated reasoning 
methods~\cite{holyhammer}, modified to learn from multiple libraries and able
to predict advice based on multiple libraries. In this research 
 we will focus on sharing proof knowledge between libraries
of proof assistants based on higher-order logic, in particular \holfour~\cite{hol4}
and \hollight~\cite{Harrison09hollight}. 
Extending the approach to learning from
developments in provers that do not share the same logic lies beyond the scope of
this paper.

Once a sufficient number of matching concepts is discovered, theorems and proofs about
these concepts can be found in both libraries, and we can start to implement methods for using
the combined knowledge in future proofs. To this end, we will use the AI-ATP system \holyhammer~\cite{holyhammer}.
We will propose various scenarios augmenting the learning and prediction phases of \holyhammer to
make use of the combined proof library. In order to evaluate the approach, we will simulate
incrementally reproving a prover's library given the knowledge of the library of the other prover.
The use of the combined knowledge significantly improves the proof advice quality
provided by \holyhammer.
Our description of the approach focuses on \hollight and \holfour, but the method
can be applied to any pair of provers for which a mapping between the logics
is known.

\subsection{Related work}
As reuse of mathematical knowledge formalizations is an important problem, it has already
been tackled in a number of ways. In the context of higher-order logic, \textsf{OpenTheory}~\cite{opentheory}
provides cross-prover packages, which allow theory sharing and simplify development. These packages
provide a high-quality standard library, but need to be developed manually. The \emph{Common HOL Platform}~\cite{commonhol}
provides a way to re-use the proof infrastructure across HOL provers.


Theory morphisms provide a versatile way to prove properties of objects of the same structure. 
The idea has been tried across Isabelle formalizations in the AWE framework by
Bortin et al.~\cite{bortin06awe}. It also serves as a basis for the
\textsf{MMT} (Module system for Mathematical Theories) framework~\cite{rabe13mmt}.

With our method, this principle was developed in both directions. We first search for similar properties of 
structures to find possible morphism between different fields.
We then use these conjectured morphisms to translate the properties between the 
two fields. Our main idea is that we don't prove the isomorphism 
which is often a complex problem but we learn from the knowledge 
gained from the derived properties.  
Moreover, even when the two fields are not completely isomorphic, 
the method often gives good advice. Indeed, suppose the set of 
reals in one library were incorrectly matched to the set of 
rationals in 
the other, we can still rely on properties of rationals that are 
also true for reals.

A direct approach is to create translations between formal libraries. This can only be applied when the defined concepts have the same or equivalent definitions.
The \textsf{HOL/Import} translation from \holfour and \hollight to \isabellehol
implemented by Obua and Skalberg~\cite{obuaimport} already mapped a number of concepts.
This was further extended by the second author~\cite{ckak-itp13} to map 70 concepts,
including differently defined real numbers. \hollight has also been translated into
\coq by Keller and Werner~\cite{holcoq}. It is the first translation between systems
based on significantly different logics.
In each of these imports, the mapping of the concepts has been done manually.

Compared with manually defined translations, our approach can find the mappings and
the knowledge that is shared automatically. It can also be used to prove statements
that are slightly different and in some cases even more general.
Additionally, the proof can use preexisting theorems in the target library.
On the other hand, when a correct translation is found by hand, it is
guaranteed to succeed, while our approach relies on AI-ATP methods which fail for
some goals. The possibility of combining the two approaches is left open.

\subsubsection*{Overview}

The rest of this paper is organized as follows. In Section~\ref{sec:requirements}, we introduce the
AI-ATP system \holyhammer and describe automatic recognition of similar concepts in different
formal proof developments. In Section~\ref{sec:scenarios}, we propose
a number of scenarios for combining the knowledge of multiple provers.
In Section~\ref{sec:evaluation}, we evaluate the ability to reprove the \holfour and \hollight
libraries using the combined knowledge.
In Section~\ref{sec:concl} we conclude and present an outlook on the future work.

\section{Preliminaries}\label{sec:requirements}
\subsection{\holyhammer}\label{sec:hh}
\holyhammer~\cite{ckju-mcs-hh} is an AI-ATP proof advice system for \hollight and \holfour. Given a user conjecture, it uses machine learning to select a subset of the
accessible facts in the library, that are likely to prove the conjecture. It then translates the
conjecture together with the selected facts to
the input language of one of the available ATP systems to find the exact dependencies necessary
to prove the theorem in higher-order logic. This method is also followed by the system \sledgehammer~\cite{sledgehammer10}.

In this section we shortly describe how \holyhammer processes conjectures, as we will
augment some of these steps in Section~\ref{sec:scenarios}.
First, we describe how libraries are exported. Then, we explain 
how the exported objects and dependencies are processed to find suitable lemmas.
Finally, we briefly show how the conjecture can be proven from these lemmas. More
detailed descriptions of these steps are presented in~\cite{holyhammer,TGCK-CPP15}.

\subsubsection{Export}
We will associate each ITP library with the set of constants and theorems that it contains. In particular, the type constructors
will also be regarded as constants in this paper.
As a first step, we define a format for representing formulas in type theory, as we aim to support formulas from various provers. A subset of this format is chosen to represent the higher-order logic 
statements in \hollight and \holfour. Each object is exported in this format with 
additional information about the theory where it was created. The theory information will 
let us export incompatible developments (i.e. ones that can not be loaded into the same ITP session or even originate from different ITPs) into 
\holyhammer~\cite{ckfr-cicm14}. Additionally, we can fully preserve the names of the original constants in the 
export. Finally, the dependencies of each theorem (i.e the set of theorems which were directly 
used to proved it) are extracted. This last step is achieved by patching the kernels of
\holfour and \hollight.

\subsubsection{Premise selection}
The premise selection algorithm takes as input an (often large) set of accessible theorems, a conjecture, and the information about previous successful proofs. It returns a subset of the theorems that is likely to prove the conjecture.
It involves three phases: feature extraction, learning, and prediction.

The features of a formula are a set of characteristics of the theorem, which we represent by strings.
Depending on the choice of characterization, it can simply be the list of the constants and types
present in the formula, or the string representation of the normalized sub-terms of the formula, or
even features based on formula semantics~\cite{ckjujv-ijcai15}.
The \emph{feature extraction} algorithm takes a formula as input and computes this set.

A relation between the features of conjectures and their dependencies is inferred from the
features of all proved theorems and their dependencies by the \emph{learning} algorithm.
This step effectively finds a function that given conjecture characteristics finds the
premises that are likely to be useful to prove this conjecture. \emph{Prediction} refers to
the evaluation of this function on a given conjecture.

These phases will be influenced by the concept matching (see Section~\ref{sec:matching}) and
differentiated in each of the scenarios (see Section~\ref{sec:scenarios}).

\subsubsection{Translation and reconstruction}

A fixed number of most relevant predicted lemmas (all the experiments in this paper
fix this number to 128, as it has given best results for HOL in combination with 
E-prover~\cite{TGCK-CPP15}) are translated together
with the conjecture to an ATP problem. If an ATP prover is able to find a proof, various
reconstruction methods are attempted. The most basic reconstruction method is to inspect the
ATP proof for the premises that were necessary to prove the conjecture. This set is usually
sufficiently small, so that certified ITP proof methods (such as MESON~\cite{meson} or
Metis~\cite{metis}) can prove
the higher-order counterpart of the statement and obtain an ITP theorem.

\subsection{Concept Matching}\label{sec:matching}

{Concept matching}~\cite{tgck-cicm14} allows the automatic discovery of concepts from one proof library or proof assistant in another.
An AI-ATP method can benefit from the library combination
only when some of the concepts in the two libraries are related: Without such mappings the
sets of features of the theorems in each library are disjoint and premise selection can only
return lemmas from the library the conjecture was stated in. As more similar concepts are
matched (for example we conjecture that the type of integers in \holfour
\texttt{h4/int} and the type of integers in \hollight \texttt{hl/int} describe the same type),
the feature 
extraction mechanism will characterize theorems talking about the  matched
concepts by the same features. As a consequence, we will also get predicted lemmas from the other
library. We will discuss how such theorems from a different library can be used without
sacrificing soundness in Section~\ref{sec:scenarios}.

For a step by step of the concept matching algorithm, 
we will refer to our previous work~\cite{tgck-cicm14} and only 
present here a short summary and the changes that improve the matching for the scenarios proposed in this paper.
Our algorithm is implemented for \holfour and \hollight, but we believe the procedure can work for any pair of provers based on similar logics such as \coq~\cite{coq14} and 
\matita~\cite{Matita14}.

\subsubsection{Summary}
Our matching algorithm is based on the properties (such as associativity, commutativity, nilpotence, \ldots) of the objects 
of our logic (constants and types). If 
two objects from two libraries share a large enough number of 
relevant properties, they will eventually be matched, even though 
they may have been defined or represented 
differently.
In the description of the procedure, we will consider every type 
as a constant.
Initially, the set of matched constants contains only logical constants.
First, we give a highest weight for rare properties with a lot of 
already matched constants.
Second, we look at all possible pairs of constants and find their 
shared properties. The final score for a pair of constant is the sum of their weights amortised by the total number of properties of each constant.
The two constants with the highest similarity score are matched. 
The previous two steps are repeated until there are no more shared properties between unmatched constants.

\subsubsection{Improvements and Limitations}

The similarity scoring heuristic can be evaluated more efficiently than the ones presented in~\cite{tgck-cicm14} and is able to map more constants correctly: Thanks to a better representation of the data
the time taken to run our implementation of the matching algorithm on the standard library of 
\hollight (including complex and multivariate) and the standard library of \holfour was decreased 
from 1 hour to 5 minutes.
By computing only the initial property frequencies and using them together with 
the proportion of matched constants to influence the weight of each property in the iterative part
the time can be further decreased to 2 minutes.
The algorithm now returns 220 correct matches instead of the 178 previously 
obtained and 15 false positives (pairs that are matched but do not represent the same 
concept) instead of 32. The better results are a consequence of the inclusion of types in the 
properties and the updated scoring function.

The proposed approach can only match objects that have the same structure. In the case of the
two proof assistants we focus on, it can successfully match the types of natural numbers,
integers or real number, however it is not
able to match the dedicated \hollight type \texttt{hl/complex} to the
complex numbers of \holfour represented by pairs of real numbers \texttt{h4/pair}(\texttt{h4/real},\texttt{h4/real}).
This issue could be partially solved by the introduction of a matching between 
sub-terms combined with a directed matching. The type \texttt{hl/complex} could then be considered as pair 
of reals in \holfour. For the reverse direction, we would need to know if the pair of reals was 
intended to represent a pair of reals or a complex. One idea to solve this problem could be to create a matching 
substitution that also depends on the theorems. These general ideas could form a basis for 
a future extension of 
the matching algorithm.

\section{Scenarios}\label{sec:scenarios}

In this section we propose 
four ways an AI-ATP system can benefit from the knowledge contained in
a library of a different prover. We will call these methods ``scenarios'' and we
will call the library of a different prover ``external''. All four scenarios
require the base libraries to already be matched.
This means, that we have already computed a matching substitution from the theorems of both libraries
and in all the already available facts in the libraries, the matched constants are replaced by their
common representatives.

Throughout our scenarios, we will rely on the notion of equivalent theorems to map lemmas from one library to the other. This notion is defined below, as well as some useful notations.

\begin{definition} [Equivalent theorems]
  Two theorems are considered equivalent if their conjunctive normal forms are equal modulo the
  order of conjuncts, disjuncts, and symmetry of equality.  Given a theorem $t$, the set of the theorems equivalent to it in the library $lib$ 
  will be noted $E(lib,t)$. 
\end{definition}

\begin{remark}
This definition only makes sense if the two libraries can be represented in the same logic. This is
straightforward if the two share the same logic.
\end{remark}

\begin{definition} [Notations]
\\Given a library, we define the following notations:
\begin{itemize}
\item $Dep(t)$ stands for the set of lemmas from which a theorem $t$ was proved. We call them the dependencies of $t$. This definition is not recursive, i.e. the set does not include theorems used to prove these lemmas.
\item The function $Learn()$ infers a relation between conjectures and sets of relevant lemmas from
the relation between theorems and their dependencies. 
\item $Pred(c,L)$ is the set of lemmas related to a conjecture $c$ predicted by the relation $L$.
\end{itemize} 
\end{definition}

In each scenario, each library plays an asymmetric role. In the 
following, the library where we 
want to prove the conjecture, is called the internal or the 
initial library. In contrast, the 
library from which we get extra advice from, is called the 
external library.
In this context, using \holyhammer alone without any knowledge 
sharing is our default scenario, 
naturally named ``internal predictions''.
We illustrate each selection method by giving an example 
of a theorem that could only be reproved by its strategy. These 
examples are extracted from our experiments 
described in Section \ref{sec:evaluation}.

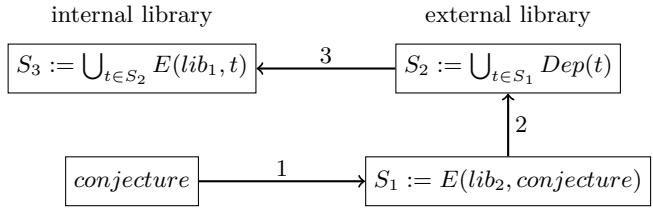
\begin{figure}[htb]
\centering
\begin{tikzpicture}[]
\node [rect] (1) {$conjecture$};
\node [rect, right of=1, node distance = 5cm] (2) {$S_1 := E(lib_2,conjecture)$};
\node [rect, above of=2, node distance = 1.5cm] (a2) {$S_2 := \bigcup_{t \in S_1} Dep(t) $};
\node [rect, above of=1, node distance = 1.5cm] (a1) {$S_3 := \bigcup_{t \in S_2} E(lib_1,t) $};
\draw[-to,black,thick] (1) -- node[yshift=5]{1} (2);
\draw[-to,black,thick] (2) -- node[xshift=5]{2} (a2);
\draw[-to,black,thick] (a2) -- node[yshift=5]{3} (a1);
\node [label=internal library, fit=(1) (a1)] (n1) {}; 
\node [label=external library, fit=(2) (a2)] (n2) {}; 

\end{tikzpicture}
\caption{Finding lemmas from dependencies in the external library.}
\label{fig:ext_dep}
\end{figure}

\subsubsection*{Scenario 1: External Dependencies} The first scenario assumes that the proof libraries
are almost identical. We compute the set of theorems equivalent to the conjecture in the
external library. For all of their dependencies, we return the lemmas in the library equivalent
to these dependencies. The scenario is presented in Fig~\ref{fig:ext_dep}.
This scenario would work very well, if the corresponding theorem is present in the external
library and a sufficient corresponding subset of its dependencies is already present in the initial library.
As this is often not the case (see Section~\ref{sec:evaluation}), we will use an AI-ATP method next.

\begin{example}
The theorem \texttt{REAL\_SUP\_UBOUND} in \holfour asserts that each element of a bounded subset of 
reals is less than its supremum. The equivalent theorem in \hollight has 3 dependencies: 
the relation between $<$ and $\leq$ \texttt{REAL\_NOT\_LT}, the antisymmetry of $<$ \texttt{REAL\_LT\_REFL} and 
the definition of supremum \texttt{REAL\_SUP}.
Each of them have one equivalent in \holfour. The resulting problem was translated and solved by an ATP and the 3 lemmas appeared in the proof.
\end{example}

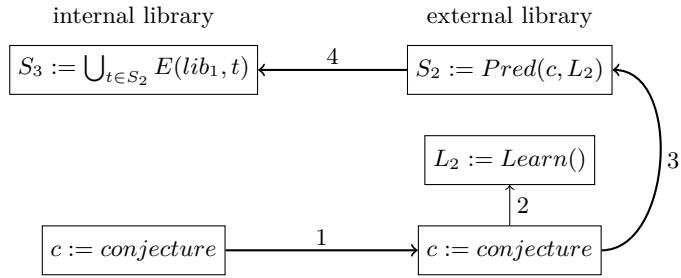
\begin{figure}[htb]
\centering
\begin{tikzpicture}[]
\node [rect] (1) {$c := conjecture$};
\node [rect, right of=1, node distance = 5cm] (2) {$c := conjecture$};
\node [rect, above of=2, node distance = 1.2cm] (a2) {$L_2 := Learn()$};
\node [rect, above of=a2, node distance = 1.2cm] (aa2) {$S_2 := Pred(c,L_2) $};
\node [rect, above of=1, node distance = 2.4cm] (a1) {$S_3 := \bigcup_{t \in S_2} E(lib_1,t) $};
\draw[-to,black,thick] (1) -- node[yshift=5]{1} (2);
\draw[-to,black] (2) -- node[xshift=5]{2} (a2);
\draw[-to,black,thick] (2) to [out=0,in=0] node[xshift=5]{3} (aa2);
\draw[-to,black,thick] (aa2) -- node[yshift=5]{4} (a1);
\node [label=internal library, fit=(1) (a1)] (n1) {}; 
\node [label=external library, fit=(2) (a2) (aa2)] (n2) {}; 

\end{tikzpicture}
\caption{Learning and predicting lemmas in the external library}
\label{fig:ext_pred}
\end{figure}

\subsubsection*{Scenario 2: External Predictions}
The next scenario is depicted in Fig~\ref{fig:ext_pred}. The steps are as follows:
We translate the conjecture to the external library (step 1). We predict
the relevant lemmas in the external library (steps 2 and 3). We map the
predicted lemmas back to the initial library using their equivalents (step 4).
To sum up, this scenario proposes an automatic way of proving a conjecture providing that the 
external library contains relevant lemmas that have equivalents in the internal library. 
One advantage of this scenario over the standard ``internal predictions'' is that the relation 
between features and dependencies is fully developed in the external library, yielding better predictions.

In our experiments, the translation step is not needed 
because the matching is already applied and the logic of 
our provers are the same.

\begin{example}
The theorem \texttt{LENGTH\_FRONT} from the \holfour theory \texttt{rich\_list} states that the 
length of a non-empty list without its last element is equal to its length minus one.
The subset of predicted lemmas used by the ATP were 6 theorems 
about natural numbers and 6 theorems about list. These theorems 
are \holfour equivalents of selected \hollight lemmas. 
\end{example}

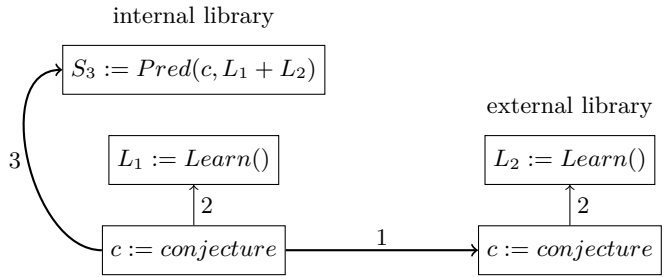
\begin{figure}[htb]
\centering
\begin{tikzpicture}[]
\node [rect] (1) {$c := conjecture$};
\node [rect, right of=1, node distance = 5cm] (2) {$c := conjecture$};
\node [rect, above of=2, node distance = 1.2cm] (a2) {$L_2 := Learn()$};
\node [rect, above of=1, node distance = 1.2cm] (a1) {$L_1 := Learn()$};
\node [rect, above of=a1, node distance = 1.2cm] (aa1) {$S_3 := Pred(c,L_1 + L_2)$};
\draw[-to,black,thick] (1) -- node[yshift=5]{1} (2);
\draw[-to,black] (2) -- node[xshift=5]{2} (a2);
\draw[-to,black] (1) -- node[xshift=5]{2} (a1);
\draw[-to,black,thick] (1) to [out=180,in=180] node[xshift=-5]{3} (aa1);
\node [label=internal library, fit=(1) (a1) (aa1)] (n1) {}; 
\node [label=external library, fit=(2) (a2)] (n2) {}; 
\end{tikzpicture}
\caption{Learning in both libraries and predicting lemmas in the internal library.}
\label{fig:int_pred_all_learn}
\end{figure}

\subsubsection*{Scenario 3: Combined Learning}
In this and the next scenario we will combine the knowledge from the external library with the
information already present in the internal library.
The scenario is presented in Fig~\ref{fig:int_pred_all_learn}.
First, the conjecture is translated to the external prover.
Second, the features suitable for proving the conjecture are learned 
from the dependencies between the theorems in both systems. 
Third, lemmas from the original library containing these features 
are predicted. In a nutshell, this scenario defines an automatic 
method, that enhances the standard ``internal predictions'' by 
including advice from the external library about the relevance of 
each feature.

\begin{example}
This example and the next one are using advice from \holfour in \hollight which means that the 
roles of the two provers are reversed compared to the first two examples.
The \hollight theorem \texttt{SQRT\_DIV} asserts that the square root of the quotient of two 
non-negative reals is equal to the quotient of their square roots. In this scenario no external 
theorems are translated but learning form the \holfour proofs still improved the 
predictions directly made in \hollight. The proof found for this theorem is based on the dual 
theorems for multiplication \texttt{SQRT\_MUL} and inversion \texttt{SQRT\_INV} and basic 
properties of division \texttt{real\_div}, multiplication \texttt{REAL\_MUL\_SYM}, inversion 
\texttt{REAL\_LE\_INV\_EQ} and absolute value \texttt{REAL\_ABS\_REFL}.
\end{example}

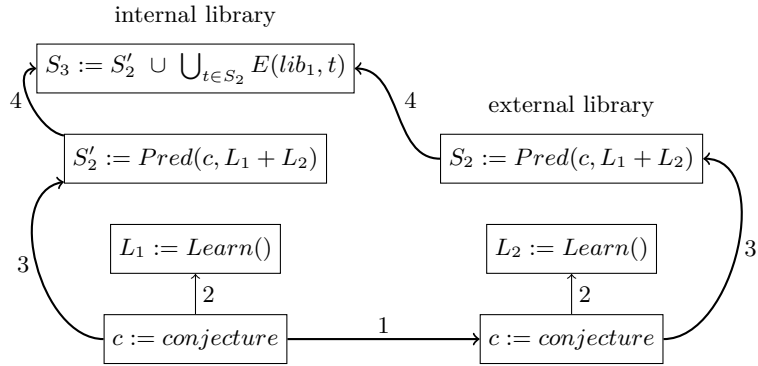
\begin{figure}[htb]
\centering
\begin{tikzpicture}[]
\node [rect] (1) {$c := conjecture$};
\node [rect, right of=1, node distance = 5cm] (2) {$c := conjecture$};
\node [rect, above of=2, node distance = 1.2cm] (a2) {$L_2 := Learn()$};
\node [rect, above of=a2, node distance = 1.2cm] (aa2) {$S_2 := Pred(c,L_1 + L_2)$};
\node [rect, above of=1, node distance = 1.2cm] (a1) {$L_1 := Learn()$};
\node [rect, above of=a1, node distance = 1.2cm] (aa1) {$S_2' := Pred(c,L_1 + L_2)$};
\node [rect, above of=aa1, node distance = 1.2cm] (aaa1) {$S_3 := S_2'\ \cup \ \bigcup_{t \in S_2} E(lib_1,t)$};
\draw[-to,black,thick] (1) -- node[yshift=5]{1} (2);
\draw[-to,black] (2) -- node[xshift=5]{2} (a2);
\draw[-to,black,thick] (2) to [out=0,in=0] node[xshift=5]{3} (aa2);
\draw[-to,black] (1) -- node[xshift=5]{2} (a1);
\draw[-to,black,thick] (1) to [out=180,in=190] node[xshift=-5]{3} (aa1);
\draw[-to,black,thick] (aa2) to [out=180,in=0] node[yshift=5,xshift=5]{4} (aaa1);
\draw[-to,black,thick] (aa1) to [out=170,in=180] node[xshift=-5]{4} (aaa1);

\node [label=internal library, fit=(1) (a1) (aa1) (aaa1)] (n1) {}; 
\node [label=external library, fit=(2) (a2) (aa2) ] (n2) {}; 
\end{tikzpicture}
\caption{Learning and predicting lemmas from both libraries.}
\label{fig:all_pred_all_learn}
\end{figure}

\subsubsection*{Scenario 4: Combined Predictions}
The last and most developed scenario, shown in Fig~\ref{fig:all_pred_all_learn},  associate 
the strategies from the two preceding scenarios, effectively learning and predicting lemmas from both
libraries. The first and second steps are the same as in ``combined learning''. The third step 
predicts lemmas in both libraries from the whole learned data. Finally, we map back the external 
predictions and return them together with the internal predictions.

\begin{example}
Let $n,m,p$ be natural numbers.\\
The \hollight theorem \texttt{HAS\_SIZE\_DIFF} declares that if a set $A$ has $n$ elements
and $B$ is a subset of A that has $m$ elements then the difference $B \setminus A$ has $n - m$ elements.
The first two lemmas necessary for the proof were directly found in \hollight. One is the definition of the constant
\texttt{HAS\_SIZE} which asserts that a set has size $p$ if and only if it is finite and has cardinality $p$.
The other \texttt{CARD\_DIFF} is almost the same as the theorem to be proved but stated for the cardinality of finite sets.
The missing piece \texttt{FINITE\_DIFF} is predicted inside the \holfour library. Its equivalent in \hollight declares that the difference of two finite sets is a finite set,
which allows the ATP to conclude.
\end{example}

\subsection{Unchecked scenarios}\label{sec:soundess}

In each of the previous scenarios, the final predicted lemmas come from the initial library. This means
that our approach is sound with respect to the internal prover. The application of the
matching substitution on one library renames the constants in all theorems injectively because no non-trivial matching is performed between two constants of the same library.

We will now consider the possibility of returning matched lemmas from the external 
library even if they do not have an equivalent in the internal one.
This means giving advice to the user in the form: ``your conjecture can be
proved using the theorems $th_1$ and $th_2$ that you already have and an
additional hypothesis with the given statement which you should be able to prove.''
To verify that these scenarios are well-founded, a user would need to prove the proposed hypotheses.
That could be achieved by either importing the theorems or applying the approach recursively.
If a constant contained in these lemmas is matched inconsistently then each method 
would fail to reprove the lemmas, preserving the coherence of the internal library.
We do not yet have an import mechanism from \holfour to \hollight (and conversely) or a recursive 
mechanism for our scenarios. In this recursive approaches, the predicted facts in 
the external library should be restricted to those proved before the conjecture 
when it has an equivalent in the external library. 
Otherwise, a loop in the recursive algorithm may be created.

We will still evaluate the ``unchecked'' scenarios to see what is the maximum
added value such mechanisms could generate.

\section{Evaluation}\label{sec:evaluation}

We perform all the experiments on a subset of the standard libraries of \hollight and \holfour.
The \holfour dataset includes 15 type constructors, 509 constants, and 3935 theorems. The
\hollight dataset contains 21 type constructors, 359 constants and 4213 theorems.
The subsets were chosen to include a variety of fields ranging from list to real analysis. The most similar pairs of theories are listed by their number of common equivalent classes of theorems in Table~\ref{tab:sim_theory}. The number of theorems in each theory is indicated in parenthesis. 

\begin{table}[htb]
\centering
\begin{tabular}{@{}ccccc@{}}
\toprule
 \holfour theory && \hollight theory && common theorems \\
\midrule
 \texttt{pred\_set}(434) && \texttt{sets}(490) && 128 \\
 \texttt{real}(469) && \texttt{real}(291) && 81 \\
 \texttt{poly}(87) && \texttt{poly}(142) && 72 \\
 \texttt{bool}(177) && \texttt{theorems}(90) && 61\\
 \texttt{transc}(229) && \texttt{transc}(355) && 58\\ 
 \texttt{arithmetic}(385) && \texttt{arith}(245) && 57\\
 \texttt{integral}(83) && \texttt{transc}(355) && 48 \\ 
 \bottomrule
\end{tabular}
\caption{The seven most similar pairs of theories by their number of common equivalent classes of theorems according to our matching}
\label{tab:sim_theory}
\end{table}

The matching, predictions, and the preparation of the ATP problems have been done on a
laptop with 4 Intel Core i5-3230M 2.60GHz processors and 3.6 GB RAM. All ATP problems
are evaluated on a server with 48 AMD Opteron 6174 2.2 GHz CPUs, 320 GB RAM and 0.5 MB L2
cache per CPU. A single core is assigned to each ATP problem. The ATP used is \eprover version
1.8 running in the automatic mode with a time limit of 30 seconds.

\subsubsection{Simulation}
We will try to prove each theorem in an environment, where information is restricted to
the one that was available when this theorem was proved. This amounts to:
\begin{itemize}
\item forgetting that it is a theorem and the knowledge of its dependencies,
\item finding the subset of facts in the library that are accessible from this theorem,
\item computing the matching with the other library based on this subset only,
\item predicting lemmas from this subset (plus the other library in the ``unchecked'' scenarios).
\end{itemize}

For the purpose of our simulation, the external library is always completely known, as we
suppose that it was created previously. In reality, the two libraries were developed
in parallel, with many \holfour theories available before similar
formalizations in \hollight have been performed.

In Fig.~\ref{fig:matching_history}, we show the evolution of the number of matched constants and
compare it to the number of declared constants in the theory during the incremental reproving of 
two theories. The first graph shows that the number of matched constants stagnate whereas the 
declared constants continue to increase in the second half of the theory. This suggests that theories
formalizing the same concepts may be developed in different directions for each prover. The second 
graph indicates a better coverage of the \hollight theory \texttt{lists}. In the beginning, the 
number of matched constants grows even more rapidly than the number of declared constants because new 
matches are found for constants defined in previous theories.

\begin{center}
\begin{figure}[]
\begin{center}
\begin{tikzpicture}[]
  \begin{axis}[ 
            height=5cm,
            width=6cm,
            xmin=-5, xmax=431,
            ymin=0, ymax=80,
            xtick={},
            ytick={10,20,30,40,50,60,70},
            ylabel={New constants}
          ]  
\addplot [green] coordinates{      
(0,0)(1,0)(2,0)(3,0)(4,0)(5,0)(6,7)(7,7)(8,7)(9,10)(10,10)(11,10)(12,10)(13,10)(14,10)(15,10)(16,10)(17,10)(18,10)(19,10)(20,10)(21,10)(22,10)(23,10)(24,10)(25,10)(26,10)(27,10)(28,10)(29,10)(30,10)(31,10)(32,10)(33,11)(34,11)(35,11)(36,11)(37,11)(38,11)(39,11)(40,11)(41,11)(42,11)(43,11)(44,11)(45,11)(46,11)(47,11)(48,11)(49,11)(50,11)(51,11)(52,11)(53,11)(54,11)(55,11)(56,11)(57,11)(58,11)(59,11)(60,11)(61,11)(62,11)(63,11)(64,11)(65,11)(66,11)(67,11)(68,11)(69,11)(70,11)(71,11)(72,11)(73,11)(74,11)(75,11)(76,11)(77,11)(78,11)(79,11)(80,11)(81,11)(82,11)(83,11)(84,11)(85,11)(86,11)(87,11)(88,11)(89,11)(90,11)(91,11)(92,11)(93,11)(94,11)(95,11)(96,11)(97,11)(98,11)(99,11)(100,11)(101,11)(102,11)(103,11)(104,11)(105,11)(106,11)(107,11)(108,11)(109,11)(110,11)(111,11)(112,11)(113,11)(114,11)(115,11)(116,11)(117,11)(118,11)(119,11)(120,11)(121,11)(122,11)(123,11)(124,11)(125,11)(126,11)(127,11)(128,11)(129,11)(130,11)(131,11)(132,11)(133,11)(134,11)(135,11)(136,11)(137,11)(138,11)(139,11)(140,11)(141,11)(142,11)(143,11)(144,11)(145,11)(146,11)(147,11)(148,11)(149,11)(150,11)(151,11)(152,11)(153,11)(154,11)(155,11)(156,11)(157,11)(158,11)(159,11)(160,11)(161,11)(162,11)(163,11)(164,11)(165,11)(166,11)(167,11)(168,12)(169,12)(170,12)(171,12)(172,12)(173,12)(174,12)(175,12)(176,14)(177,14)(178,14)(179,14)(180,14)(181,14)(182,14)(183,14)(184,14)(185,14)(186,14)(187,14)(188,14)(189,14)(190,14)(191,14)(192,14)(193,14)(194,14)(195,14)(196,14)(197,14)(198,14)(199,14)(200,14)(201,14)(202,14)(203,14)(204,14)(205,14)(206,14)(207,14)(208,14)(209,14)(210,14)(211,14)(212,14)(213,14)(214,14)(215,14)(216,14)(217,14)(218,14)(219,14)(220,14)(221,14)(222,14)(223,14)(224,14)(225,14)(226,14)(227,14)(228,14)(229,14)(230,14)(231,14)(232,14)(233,14)(234,14)(235,14)(236,14)(237,14)(238,14)(239,15)(240,15)(241,15)(242,15)(243,15)(244,15)(245,15)(246,15)(247,15)(248,15)(249,15)(250,15)(251,15)(252,15)(253,15)(254,15)(255,15)(256,15)(257,15)(258,15)(259,15)(260,15)(261,15)(262,15)(263,15)(264,15)(265,15)(266,15)(267,15)(268,15)(269,15)(270,15)(271,15)(272,15)(273,15)(274,15)(275,15)(276,15)(277,15)(278,15)(279,15)(280,15)(281,15)(282,15)(283,15)(284,15)(285,15)(286,15)(287,15)(288,15)(289,15)(290,15)(291,15)(292,15)(293,15)(294,15)(295,15)(296,15)(297,15)(298,15)(299,15)(300,15)(301,15)(302,15)(303,15)(304,15)(305,15)(306,15)(307,15)(308,15)(309,15)(310,15)(311,15)(312,15)(313,15)(314,15)(315,15)(316,15)(317,15)(318,15)(319,15)(320,15)(321,15)(322,15)(323,15)(324,15)(325,15)(326,15)(327,15)(328,15)(329,15)(330,15)(331,15)(332,15)(333,15)(334,15)(335,15)(336,15)(337,15)(338,15)(339,15)(340,15)(341,15)(342,15)(343,15)(344,15)(345,15)(346,15)(347,15)(348,15)(349,15)(350,15)(351,15)(352,15)(353,15)(354,15)(355,15)(356,15)(357,15)(358,15)(359,15)(360,15)(361,15)(362,15)(363,15)(364,15)(365,15)(366,15)(367,15)(368,15)(369,15)(370,15)(371,15)(372,15)(373,15)(374,15)(375,15)(376,15)(377,15)(378,15)(379,15)(380,15)(381,15)(382,15)(383,15)(384,15)(385,15)(386,15)(387,15)(388,15)(389,15)(390,15)(391,15)(392,15)(393,15)(394,15)(395,15)(396,15)(397,15)(398,15)(399,15)(400,15)(401,15)(402,15)(403,15)(404,15)(405,15)(406,15)(407,15)(408,15)(409,15)(410,15)(411,15)(412,15)(413,15)(414,15)(415,15)(416,15)(417,15)(418,15)(419,15)(420,15)(421,15)(422,15)(423,15)(424,15)(425,15)(426,15)(427,15)(428,15)(429,15)(430,15)
 };
\addplot [dashed, blue] coordinates{      
(0,0)(1,0)(2,0)(3,0)(4,0)(5,0)(6,0)(7,9)(8,9)(9,9)(10,15)(11,15)(12,16)(13,16)(14,16)(15,16)(16,19)(17,19)(18,19)(19,19)(20,19)(21,19)(22,19)(23,19)(24,19)(25,19)(26,19)(27,19)(28,19)(29,19)(30,19)(31,19)(32,19)(33,19)(34,19)(35,19)(36,19)(37,19)(38,19)(39,19)(40,19)(41,19)(42,19)(43,19)(44,19)(45,19)(46,19)(47,19)(48,19)(49,19)(50,19)(51,19)(52,19)(53,19)(54,19)(55,19)(56,19)(57,19)(58,19)(59,19)(60,19)(61,19)(62,19)(63,19)(64,19)(65,19)(66,19)(67,19)(68,19)(69,19)(70,19)(71,19)(72,19)(73,19)(74,19)(75,19)(76,19)(77,19)(78,19)(79,19)(80,19)(81,19)(82,19)(83,19)(84,19)(85,19)(86,19)(87,19)(88,19)(89,19)(90,19)(91,19)(92,19)(93,19)(94,19)(95,19)(96,19)(97,19)(98,19)(99,19)(100,19)(101,19)(102,19)(103,19)(104,19)(105,19)(106,19)(107,19)(108,19)(109,19)(110,19)(111,19)(112,19)(113,19)(114,19)(115,19)(116,19)(117,19)(118,19)(119,19)(120,19)(121,19)(122,19)(123,19)(124,19)(125,19)(126,20)(127,20)(128,20)(129,20)(130,20)(131,20)(132,20)(133,20)(134,20)(135,20)(136,20)(137,20)(138,20)(139,20)(140,20)(141,20)(142,20)(143,20)(144,20)(145,20)(146,20)(147,20)(148,20)(149,20)(150,20)(151,20)(152,22)(153,22)(154,22)(155,22)(156,22)(157,22)(158,22)(159,22)(160,22)(161,22)(162,22)(163,22)(164,22)(165,22)(166,22)(167,22)(168,22)(169,23)(170,23)(171,23)(172,23)(173,23)(174,23)(175,23)(176,23)(177,25)(178,25)(179,25)(180,25)(181,25)(182,25)(183,25)(184,25)(185,25)(186,25)(187,25)(188,25)(189,27)(190,27)(191,27)(192,27)(193,27)(194,27)(195,27)(196,27)(197,27)(198,27)(199,27)(200,27)(201,28)(202,28)(203,28)(204,28)(205,28)(206,28)(207,28)(208,28)(209,28)(210,29)(211,29)(212,29)(213,29)(214,29)(215,29)(216,29)(217,29)(218,29)(219,29)(220,30)(221,30)(222,30)(223,30)(224,30)(225,30)(226,30)(227,30)(228,30)(229,30)(230,30)(231,30)(232,30)(233,30)(234,30)(235,30)(236,30)(237,30)(238,31)(239,31)(240,31)(241,31)(242,31)(243,31)(244,31)(245,31)(246,31)(247,31)(248,32)(249,32)(250,32)(251,33)(252,33)(253,33)(254,33)(255,33)(256,33)(257,33)(258,33)(259,33)(260,33)(261,33)(262,33)(263,33)(264,33)(265,33)(266,33)(267,33)(268,33)(269,33)(270,34)(271,35)(272,35)(273,35)(274,37)(275,37)(276,37)(277,37)(278,37)(279,37)(280,37)(281,37)(282,37)(283,37)(284,37)(285,37)(286,37)(287,37)(288,37)(289,37)(290,37)(291,37)(292,37)(293,37)(294,37)(295,37)(296,37)(297,37)(298,37)(299,37)(300,37)(301,37)(302,37)(303,37)(304,39)(305,39)(306,39)(307,39)(308,40)(309,40)(310,40)(311,40)(312,40)(313,40)(314,40)(315,40)(316,40)(317,41)(318,41)(319,41)(320,41)(321,41)(322,41)(323,41)(324,41)(325,41)(326,41)(327,41)(328,41)(329,41)(330,43)(331,43)(332,43)(333,43)(334,43)(335,43)(336,44)(337,44)(338,45)(339,45)(340,45)(341,45)(342,45)(343,45)(344,45)(345,45)(346,47)(347,47)(348,47)(349,47)(350,47)(351,48)(352,48)(353,48)(354,48)(355,48)(356,48)(357,48)(358,48)(359,49)(360,49)(361,49)(362,49)(363,49)(364,49)(365,49)(366,49)(367,49)(368,49)(369,49)(370,49)(371,49)(372,49)(373,49)(374,49)(375,49)(376,49)(377,49)(378,49)(379,49)(380,49)(381,49)(382,49)(383,49)(384,49)(385,49)(386,49)(387,49)(388,49)(389,49)(390,49)(391,49)(392,49)(393,49)(394,49)(395,49)(396,49)(397,49)(398,49)(399,49)(400,49)(401,49)(402,49)(403,49)(404,49)(405,49)(406,49)(407,49)(408,49)(409,49)(410,49)(411,49)(412,50)(413,50)(414,50)(415,50)(416,50)(417,50)(418,50)(419,50)(420,50)(421,50)(422,50)(423,50)(424,50)(425,50)(426,50)(427,50)(428,50)(429,50)(430,50)(431,50)

 };
        \legend{matched constants, declared constants}
        \end{axis}
\end{tikzpicture}
\begin{tikzpicture}[]
  \begin{axis}[ 
            height=5cm,
            width=6cm,
            xmin=-2, xmax=125,
            ymin=0, ymax=50,
            xtick={},
            ytick={10,20,30,40},
          ]  
\addplot [green] coordinates{      
(0,0)(1,3)(2,5)(3,5)(4,6)(5,6)(6,6)(7,7)(8,7)(9,7)(10,8)(11,9)(12,9)(13,9)(14,10)(15,11)(16,11)(17,12)(18,12)(19,13)(20,13)(21,13)(22,13)(23,13)(24,13)(25,13)(26,13)(27,13)(28,13)(29,13)(30,13)(31,13)(32,13)(33,13)(34,13)(35,13)(36,13)(37,14)(38,14)(39,14)(40,14)(41,14)(42,14)(43,14)(44,14)(45,14)(46,14)(47,14)(48,14)(49,14)(50,14)(51,14)(52,14)(53,14)(54,14)(55,14)(56,14)(57,14)(58,14)(59,14)(60,14)(61,14)(62,14)(63,14)(64,14)(65,14)(66,14)(67,14)(68,14)(69,14)(70,14)(71,14)(72,14)(73,14)(74,14)(75,14)(76,14)(77,14)(78,14)(79,14)(80,14)(81,14)(82,14)(83,14)(84,14)(85,14)(86,14)(87,14)(88,14)(89,14)(90,14)(91,14)(92,14)(93,14)(94,14)(95,14)(96,14)(97,14)(98,14)(99,14)(100,14)(101,14)(102,14)(103,14)(104,15)(105,15)(106,15)(107,15)(108,15)(109,15)(110,15)(111,15)(112,15)(113,15)(114,15)(115,15)(116,15)(117,15)(118,15)(119,15)(120,15)(121,15)(122,15)(123,15)(124,15)
  };
\addplot [dashed, blue] coordinates{      
(0,0)(1,0)(2,1)(3,2)(4,3)(5,3)(6,4)(7,4)(8,5)(9,5)(10,6)(11,6)(12,7)(13,8)(14,8)(15,9)(16,9)(17,10)(18,10)(19,11)(20,11)(21,12)(22,12)(23,13)(24,13)(25,14)(26,14)(27,15)(28,15)(29,15)(30,15)(31,15)(32,15)(33,16)(34,16)(35,16)(36,16)(37,17)(38,17)(39,18)(40,18)(41,19)(42,20)(43,20)(44,20)(45,20)(46,21)(47,21)(48,21)(49,21)(50,21)(51,21)(52,21)(53,21)(54,21)(55,21)(56,21)(57,21)(58,21)(59,21)(60,21)(61,21)(62,21)(63,21)(64,21)(65,21)(66,21)(67,21)(68,21)(69,21)(70,21)(71,21)(72,21)(73,21)(74,21)(75,21)(76,21)(77,21)(78,21)(79,21)(80,21)(81,21)(82,21)(83,21)(84,21)(85,21)(86,21)(87,21)(88,21)(89,21)(90,21)(91,21)(92,21)(93,21)(94,21)(95,21)(96,21)(97,21)(98,21)(99,21)(100,21)(101,21)(102,21)(103,21)(104,21)(105,21)(106,21)(107,21)(108,21)(109,21)(110,21)(111,21)(112,21)(113,21)(114,21)(115,21)(116,21)(117,21)(118,21)(119,21)(120,21)(121,21)(122,21)(123,21)(124,21)(125,23)
  };
        \legend{matched constants, declared constants}
        \end{axis}
\end{tikzpicture}
\end{center}
  \caption{Evolution of the number of matched constants in the \holfour theory \texttt{list} and in the \hollight theory \texttt{lists}} 
    \label{fig:matching_history}    
    
 \end{figure}
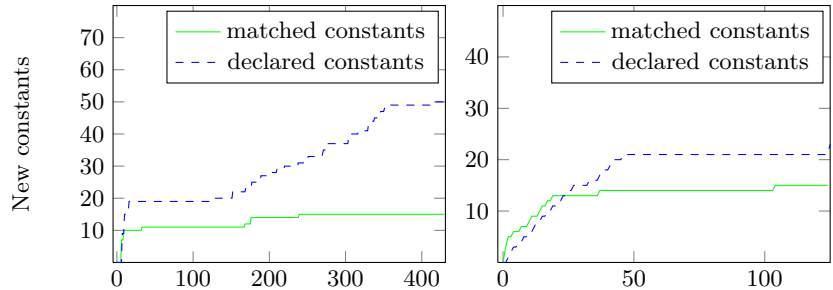 
\end{center}

\subsubsection{Results}

\begin{table}[H]
\centering
\begin{tabular}{@{}ccc@{}}
\toprule
  Scenario & checked(\%) & unchecked(\%) \\
\midrule
  empty & 4.19 & \\
  external dependencies & 5.06 (23.50) & 10.75 (49.94) \\
  external predictions & 17.49 & 34.42 \\
  external any & 18.07 & 34.74 \\
  internal predictions & \textbf{43.57} &  \\
  combined learning & 44.03 & \\
  combined predictions & 44.59 & 53.46 \\
  any & \textbf{50.06 }& 55.73 \\
  any checked or unchecked & \multicolumn{2}{c}{62.80} \\
\bottomrule
\end{tabular}
\caption{Percentage of reproved theorems in the \holfour library (internal) 
 with the knowledge from the \hollight library (external).}
\label{reproving_h4}
\end{table}

\begin{table}[H]
\centering
\begin{tabular}{@{}ccc@{}}
\toprule
  Scenario & checked(\%) & unchecked(\%)\\
\midrule
  empty  & 3.14 &  \\
  external dependencies & 6.08 (29.22) & 10.11 (48.63) \\
  external predictions & 12.74 & 33.94 \\
  external any & 13.55 & 34.32 \\
  internal predictions & \textbf{30.92} &  \\
  combined learning & 35.13 &   \\
  combined predictions & 35.56 & 44.06 \\ 
  any & \textbf{40.19} & 47.07 \\
   any checked or unchecked & \multicolumn{2}{c}{54.71} \\
\bottomrule
\end{tabular}
\caption{Percentage of reproved theorems in the \hollight library (internal) 
 with the knowledge from the \holfour library (external).}
\label{reproving_hl}
\caption*{ In the first column, scenarios are listed based on their predicted lemmas.\\
 \textbf{empty}: no lemmas \\
 \textbf{external dependencies}: dependencies of equivalent external theorems\\
 \textbf{external predictions}: external lemmas from external advice\\
 \textbf{external any}: problems solved by any of the two previous scenarios\\
 \textbf{internal predictions}: internal lemmas from internal advice\\
 \textbf{combined learning}: internal lemmas from external and internal advice\\
 \textbf{combined predictions}: external and internal lemmas from external and internal advice\\
 \textbf{any}: problems solved by at least one scenario of the same column\\
 In the second column, we restrict ourself from using external theorems that do not have an 
 internal equivalent, where as we allow it in the third column. The last line combines all the problems solved by at least one checked or unchecked scenario.
 }

\end{table}

The success rates for each scenario and each proof assistant are compiled in 
Tables~\ref{reproving_h4} and \ref{reproving_hl}.
The scenario ``empty'' gives the number of facts provable
without lemmas and is fully subsumed by the other methods.

The external dependencies scenario is the only one that is not directly comparable to the others, as 
it was performed only on the theorems that have an equivalent in the other library (876 in 
\hollight and 847 in \holfour). The percentage of theorems proved by this strategy
relative to its experimental subset is shown in parentheses. This strategy is
quite efficient on its subset but contributes weakly to the overall improvement.
These results are combined with the ``external predictions'' scenario 
to evaluate what can be reproved with external help only. 
In \holfour, the combined learning and predictions increases the 
number of problems solved over the initial  ``internal predictions'' approach
only by one percent.
The improvement is sharper in \hollight. It suggests that \holfour provides a better set 
for the learning algorithm.
The improvement provided by all scenarios can be combined to yield a significant
gain compared to the performance of \holyhammer alone, namely additional 6.5\% of all \holfour and
 9.3\% of all \hollight theorems. Another 10--15\%
could be added by the ``unchecked'' scenarios.

\subsubsection{Results by theory}\label{sec:theory_comparison}
In Table~\ref{tab:theories_h4_hl}, we investigate the performance of the 
``external dependencies'' scenario on the largest theories in our dataset. Some theories only minimally benefit from the external help. This is the case for \texttt{rich\_list} and 
\texttt{iterate}, where only few correct mappings could be found. We can see asymmetric results in pairs of similar theories. For example, the 
\texttt{real} theory in \hollight can be 72.16\% reproved from \holfour theories whereas the similar theory in \holfour does not benefit as much. This suggest that the \texttt{real} theory \holfour is more dense than its counterpart. A similar effect is observed for the \texttt{transc} formalization. The theories \texttt{pred\_set} and \texttt{sets} seem to be comparably dense.

\begin{table}[h]
\centering
\begin{tabular}{@{}ccccccc@{}}
\toprule
Scenario & \texttt{real} & \texttt{pred\_set} & \texttt{list} & \texttt{arithmetic} & \texttt{rich\_list} & \texttt{transc} \\
\midrule
external dependencies & 30.91 & 24.65 & 10.23 & 18.18 & 1.52 & 5.24 \\
\bottomrule
\\
\toprule
Scenario & \texttt{sets} & \texttt{analysis} & \texttt{transc} & \texttt{int} & \texttt{iterate} & 
\texttt{real} \\
\midrule
external dependencies & 25.51 & 27.1 & 25.91 & 52.61 & 5.47 & 72.16\\
\bottomrule
\end{tabular}
\caption{Reproving success rate in the six largest theories in 
\holfour using \hollight and the ``checked external dependencies'' 
scenario, as well as in the six largest \hollight theories using 
\holfour.}
\label{tab:theories_h4_hl}
\end{table}

\section{Conclusion}\label{sec:concl}

We proposed several methods for combining the knowledge of two ITP systems
in order to prove more theorems automatically. The methods adapt the
premise selection and proof advice components of the \holyhammer system
to include the knowledge of an external prover. In order to do it,
the concepts defined in both libraries are related through an improved 
matching algorithm. As the constants in two libraries become related, so
are the statements of the theorems. Machine learning algorithms can combine
the information about the dependencies in each library to predict useful
dependencies more accurately.

We evaluated the influence of an external library on the quality of advice,
by reproving all the theorems in a large subset of the \holfour and \hollight
standard libraries. External knowledge can improve the success from 43\% to 50\% in \holfour and from 30\% to 40\% in the number of \hollight solved goals. 
This number could reach 54\% for \holfour and
62\% for \hollight if we include the ``unchecked'' scenarios, where the
user is not only suggested known theorems, but also hypotheses left to prove.
Proving such proposed lemmas, either with the help of a translation or by
calling an AI-ATP method with shared knowledge is left as future work.


The proposed approach evaluated the influence of an external proof assistant library
for the quality of learning and prediction. An extension of the approach could be used
inside a single library: mappings of concepts inside a single library, such as those
the work of Autexier and Hutter~\cite{sergeautexier}, could provide additional
knowledge for a learning-reasoning system.

\section*{Acknowledgments}
This work has been supported by the Austrian Science Fund (FWF): P26201.

\bibliographystyle{plain}
\bibliography{biblio}

\end{document}